\documentclass[sn-mathphys, iicol]{sn-jnl}

\jyear{2021}%

\theoremstyle{thmstyleone}%
\theoremstyle{thmstyletwo}%

\theoremstyle{thmstylethree}%

\usepackage{bbding}
\usepackage{makecell}
\begin{document}

\title[An Empirical Study on Multi-Domain Robust Semantic Segmentation]{An Empirical Study on Multi-Domain Robust Semantic Segmentation}

\author[1]{\fnm{Yajie} \sur{Liu}}\email{yajie\_liu@buaa.edu.cn}
\author[2]{\fnm{Pu} \sur{Ge}}\email{pu\_ge@buaa.edu.cn}
\equalcont{These authors contributed equally to this work.}

\author*[1]{\fnm{Qingjie} \sur{Liu}}\email{qingjie.liu@buaa.edu.cn}
\author[1]{\fnm{Shichao} \sur{Fan}}\email{shichaoFan@buaa.edu.cn}
\author[1]{\fnm{Yunhong} \sur{Wang}}\email{yhwang@buaa.edu.cn}
\affil[1]{\orgdiv{State Key Laboratory of Virtual Reality Technology and Systems}, \orgname{Beihang University}, \city{Beijing}, \country{China}}

\affil[2]{\orgdiv{Hangzhou Innovation Institute}, \orgname{Beihang University}, \city{Hangzhou}, \country{China}}


\abstract{

How to effectively leverage the plentiful existing datasets to train a robust and high-performance model is of great significance for many practical applications. 
However, a model trained on a naive merge of different datasets tends to obtain poor performance due to annotation conflicts and domain divergence.
In this paper, we attempt to train a unified model that is expected to perform well across domains on several popularity segmentation datasets.
We conduct a detailed analysis of the impact on model generalization from three aspects of data augmentation, training strategies, and model capacity.
Based on the analysis, we propose a robust solution that is able to improve model generalization across domains.
Our solution ranks 2nd on RVC 2022 semantic segmentation task, with a dataset only 1/3 size of the 1st model used.
}




\maketitle

\section{Introduction}\label{sec1}

Deep learning approaches have achieved remarkable success in computer vision in recent years. Nevertheless, academic advancements that perform well on specific benchmarks often fail to be applied in real-world scenarios. With the emergence of large-scale annotated datasets, mixing datasets from diverse domains can improve model generalization in actual applications, such as autonomous driving. In addition, a unified model trained on multiple domains simplifies the decision procedure and reduces the burden on deployment. 
However, expecting a unified model performs as well as the best dedicated model in any environment is an intractable problem.

A naive combination of diverse datasets tends to  decrease the model accuracy. One main issue is that different datasets have different definitions of taxonomy. Specifically, there naively exists annotation conflicts and inconsistent category granularity due to different collection and labeling rules of datasets. Therefore, models trained on multiple datasets yield poor performance owing to a lack of compatible category supervision. Previous work~\cite{lambert2020mseg} creates a composite dataset that unifies datasets from different domains, which needs to take a tremendous annotation effort to solve category conflicts.
Besides, the dataset merged across different domains is prone to cause category imbalance problem.

Another underlying reason is the domain shift when different datasets are directly used for training a model together. To address this problem, domain generalization (DG) becomes a popular research topic with thousands of approaches being published annually~\cite{DBLP:journals/corr/GaninUAGLLML15,DBLP:conf/cvpr/IsolaZZE17,DBLP:conf/cvpr/LiPWK18,DBLP:conf/cvpr/GongLCG19,DBLP:conf/eccv/LiTGLLZT18,DBLP:conf/cvpr/ShaoLLY19,DBLP:journals/corr/abs-1909-08531,DBLP:journals/corr/TzengHZSD14,DBLP:conf/mm/WangFCYHY18}. Some studies based on domain generalization attempt to minimize the representation discrepancy between multiple domains so that the learned model can have strong generalizable and transferable capability. While learning invariant features without damaging the discriminative ability for special tasks remains a challenging problem.

In addition, approaches based on multi-domain learning~\cite{DBLP:conf/wacv/VarmaSNCJ19,DBLP:conf/iccv/BerrielLNKOS019,DBLP:conf/cvpr/XiaoG020,DBLP:conf/nips/RebuffiBV17,DBLP:journals/pami/RosenfeldT20,DBLP:journals/corr/BilenV17} also provide solutions for the joint training of multiple datasets. 
The key to multi-domain learning is to exploit correlation among multiple domains to improve the model performance with inputs from different datasets. However, these methods usually require some adaptation modules to adapt to multiple domains~\cite{DBLP:conf/nips/RebuffiBV17,DBLP:journals/pami/RosenfeldT20,DBLP:journals/corr/BilenV17}, which adds extra cost.

As explained above, the merged dataset that combines semantic segmentation datasets from different domains is complicated, with different labeling rules, unbalanced classes, and domain biases. 
Therefore, a robust model with strong representation power is required to overcome or reduce the impact of domain difference and annotation conflicts.
We believe that high model capacity can alleviate the domain shift. To further verify our hypothesis, we analyze the  capacities of different architectures and their robustness to domain shift, in particular, including Segformer~\cite{DBLP:conf/nips/XieWYAAL21} and ResNet~\cite{DBLP:conf/cvpr/HeZRS16}.
Some recent works~\cite{DBLP:journals/corr/abs-2107-13389,DBLP:journals/corr/abs-1905-04899,9709604} have shown that the “mixup” data augmentation methods contribute to improving the model robustness against the out-of-distribution performance and reducing category imbalance.
Inspired by this, we introduce and analyze three heuristic data augmentation approaches: Copy-Paste~\cite{DBLP:journals/corr/abs-2110-05474,DBLP:journals/corr/abs-2012-07177}, DomainMix~\cite{DBLP:journals/corr/abs-2107-13389} and CutMix~\cite{DBLP:journals/corr/abs-1905-04899}.
We also conduct and examine some training strategies that can further boost the model accuracy for multi-domain learning.

The main contributions are summarized as follows:
 \begin{itemize}
 \item We investigate the data augmentations, training strategies and network architectures for robust multi-domain semantic segmentation learning. Meanwhile, we apply our proposed solution to the ECCV Robust Vision Challenge 2022 (RVC) semantic segmentation task and are ranked second at a low hardware cost.
 \item We explore the impact of model capacity and model architecture on domain shift in the multi-domain learning realm. Our experiments indicate that model with strong representation capacity helps boost the generalization and robustness across datasets.
 \item We discuss the impact of domain generalization approaches on multi-domain learning task. We show that learning invariant features based on DG is prone to damage the discriminative ability.
 \end{itemize}

\section{Related Work}\label{sec2}

\textbf{Multi-domain Learning.}
The key to multi-domain learning is to learn a single model that can generalize well to all the domains. With the emergence of large-scale annotated datasets, it is of great significance to mix multiple datasets for improving model performance and adapting to a wide range of scenarios. Ros et al.~\cite{DBLP:journals/corr/RosSAW16} applied a composite dataset that combines six datasets from different domains in the driving scene to train the semantic segmentation models. Varma et al.~\cite{DBLP:conf/wacv/VarmaSNCJ19} observed that domain-cross learning leads to a decrease in model accuracy on each dataset compared to “self-training” (training on the single dataset). Lambert et al.~\cite{lambert2020mseg} proposed a consistent taxonomy that bridges datasets from multiple domains to address the incompatibility of dataset taxonomies.

To address the domain shift in multi-domain learning, Li et al.~\cite{8954111} proved that it is possible to build universal parameters that can be shared among multiple domains. Moreover, adapting to the specified domain on both early and late layers can help the model obtain the best performance. Some approaches~\cite{DBLP:conf/iccv/BerrielLNKOS019,DBLP:conf/cvpr/XiaoG020} proposed to share a common backbone architecture across multiple domains but employ a domain-specific head on each domain.
While these works usually require some adaptation components, e.g., domain-specific conv~\cite{DBLP:conf/nips/RebuffiBV17,DBLP:journals/pami/RosenfeldT20} and batch normalization~\cite{DBLP:journals/corr/BilenV17}, to adapt to different domains.
Our approach aims at learning a unified model that is adapted to diverse domains without any specific adaptation components.

\noindent\textbf{Domain Generalization.} 
The goal of Domain Generalization (DG) is to learn from data of multiple source domains and generalize well to unseen target domains, which is essentially an out-of-distribution (OOD) problem. Ben-David et al.~\cite{DBLP:conf/nips/Ben-DavidBCP06} proposed that the feature representations can be generalized to different domains when multiple source domains gain domain-invariant features. Therefore, a series of researches are presented that aim to learn domain-invariant features by reducing the representation discrepancy among different domains.

Many domain generalization approaches apply domain adversarial training for domain-invariant feature learning. Ganin et al.~\cite{DBLP:journals/corr/GaninUAGLLML15} presented a deep feed-forward network, called domain-adversarial neural network (DANN), which adds a discriminator to address the domain shift across different domains. The discriminator is trained to distinguish the domains and encourages the model to learn invariant features. Wang et al.~\cite{DBLP:journals/tmi/WangYYFH19} introduced the patch discriminator (PatchGAN)~\cite{DBLP:conf/cvpr/IsolaZZE17} for adversarial learning in medical segmentation. 
Some other methods~\cite{DBLP:conf/cvpr/LiPWK18,DBLP:conf/cvpr/GongLCG19,DBLP:conf/eccv/LiTGLLZT18,DBLP:conf/cvpr/ShaoLLY19} based on domain adversarial learning were also proposed for domain-invariant feature learning.
Another line of works for learning domain-invariant features are based on explicit feature distribution alignment. Some approaches~\cite{DBLP:journals/corr/abs-1909-08531,DBLP:journals/corr/TzengHZSD14,DBLP:conf/mm/WangFCYHY18} adopted the minimizing maximum mean discrepancy (MMD) loss to learn general representations by minimizing the distribution divergence among different domains. 

Invariant risk minimization (IRM) based approaches~\cite{https://doi.org/10.48550/arxiv.1907.02893,DBLP:conf/icml/AhujaSVD20,DBLP:conf/icml/KruegerCJ0BZPC21} do not seek to match the representation distribution of all domains, but encourage the optimal classifier on top of the representation space to be the same across all domains. IRM has recently obtained notable achievement in domain generalization. However, it tends to fail when pseudo-invariant features and geometric skews exist. To address this problem, Li et al.~\cite{DBLP:conf/aaai/LiSWZRLK022} proposed invariant information bottleneck (IIB), which focuses on minimizing invariant risks for nonlinear classifiers and reducing the impact of pseudo-invariant features.
 
Multi-domain learning for semantic segmentation also benefits from the alignment of feature representations. 
Therefore, we attempt to apply domain adversarial learning, MMD loss and IRM domain generalization approaches to improve the performance of the unified model.

\section{Datasets Analysis and Taxonomy}\label{data analysis}

The semantic segmentation contest of Robust Vision Challenge held at ECCV 2022 provides six datasets with different characteristics. 
We analyze the characteristics of the six datasets in detail, and then adopt a unified taxonomy to obtain the merged dataset for cross-domain model training.
In addition, we discuss the current challenge of semantic segmentation multi-domain learning and analyze the main characteristics of our merged dataset.

\begin{figure*}[htbp]
\centering
\includegraphics[height=9cm, width=14cm]{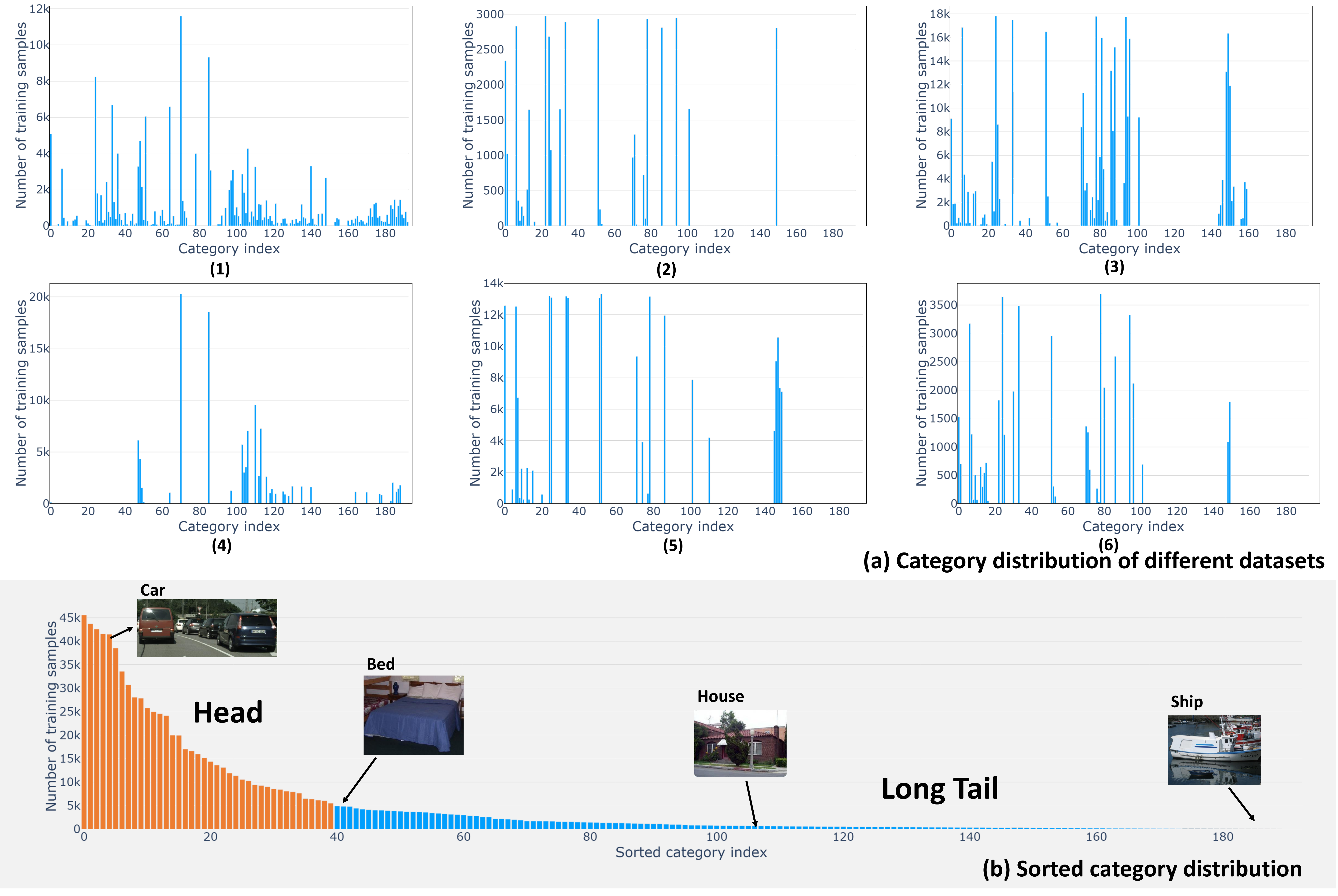}
\caption{\textbf{(a) Category distribution of different datasets.} The x-axis is the category index of 194 categories under the unified taxonomy space.
(1) ADE20K. (2) CityScapes. (3) Mapillary Vistas. (4) ScanNet. (5) VIPER. (6) WildDash.
\textbf{(b) Category distribution of the merged dataset sorted by the number of instances.} The x-axis is the category index of 194 categories sorted by the number of instances. The y-axis is the number of instances in each category.}
\label{cls_num}
\end{figure*}

\subsection{RVC 2022 Datasets}
The six datasets\footnote{All datasets used in our study are publicly available.} in the challenge are ADE20K, CityScapes, Mapillary Vistas, ScanNet, VIPER and WildDash. We introduce and list them as follows.

\textbf{ADE20K}~\cite{zhou2019semantic}, is a scene parsing dataset, which provides dense labels of 150 categories, spanning diverse annotations of scenes, objects, and objects parts. There are 20,210 images covering complex everyday scenes in the training set.  Images come from the SUN datasets~\cite{xiao2010sun} and Places~\cite{zhou2014learning}, with varying spatial resolution. The median image size is 307,200 pixels and the median aspect ratio is 4:3.

\textbf{CityScapes}~\cite{cordts2016cityscapes}, is a semantic urban scene dataset consisting of 5000 images with the same size of $1920 \times 1080$ pixels. Each image in CityScapes is with accurate pixel-wise annotation. Images were acquired from a moving vehicle during the span of several months in 50 cities, primarily in Germany. The dataset contains 2975 training images. 19 semantic labels are used for evaluation, and the void label is ignored.

\textbf{Mapillary Vistas (MVD)}~\cite{neuhold2017mapillary}, is a large-scale street-view image dataset captured over urban, countryside and off-road scenes. It is comprised of 25,000 densely-annotated images that are grouped into 66 object categories. Images are extracted from Mapillary covering mickle parts like Europe, North and South America, Asia, and Africa, and the images are diverse in terms of image resolution and aspect ratio. All images are at least FullHD, but some images are more than 22 Mpixels. The dominant aspect ratio is 4:3, and 18,000 images are used for training.

\textbf{ScanNet}~\cite{dai2017scannet}, an instance-level RGB-D dataset that includes both 2D and 3D data, contains 1513 RGB-D scans over 707 unique indoor environments. The dataset is marked in 20 classes of annotated 3D voxelized objects. The 2D semantic labels  are  acquired  from the projection of 3D aggregated annotation into its RGB-D frames, according to the computed camera trajectory. It provides a smaller subset of 25000 images, which are subsampled approximately every 100 frames from the full dataset with the size of $1296 \times 968$ pixels.

\textbf{VIPER}~\cite{richter2017playing}, a synthetic dataset, includes pixel-wise annotations of semantic and instance segmentation for 23 classes of ego\-centric driving scenes at $1920 \times 1080$ resolution. The data is collected while driving, riding, and walking 184 kilometers in a realistic virtual world by using the GTA-V game engine. The semantic segmentation training set contains 13,367 images subsampled from the video sequences.

\textbf{WildDash}~\cite{zendel2018wilddash}, a dataset for semantic segmentation for the automotive domain, includes images at $1920 \times 1080$ resolution from a variety of data sources  with many difficult scenarios (e.g. rain, road coverage, darkness, overexposure) and camera characteristics (noise, compression artifacts, distortion). Besides the classic 19 classes of Cityscapes, a total of six new labels are evaluated: van, pickup, street light, billboard, guard rail, and ego-vehicle. 4256 frames with pixel-level annotations are provided for training.

\begin{figure}[htbp]
    \centering
    \includegraphics[height=5.6cm, width=8.0cm]{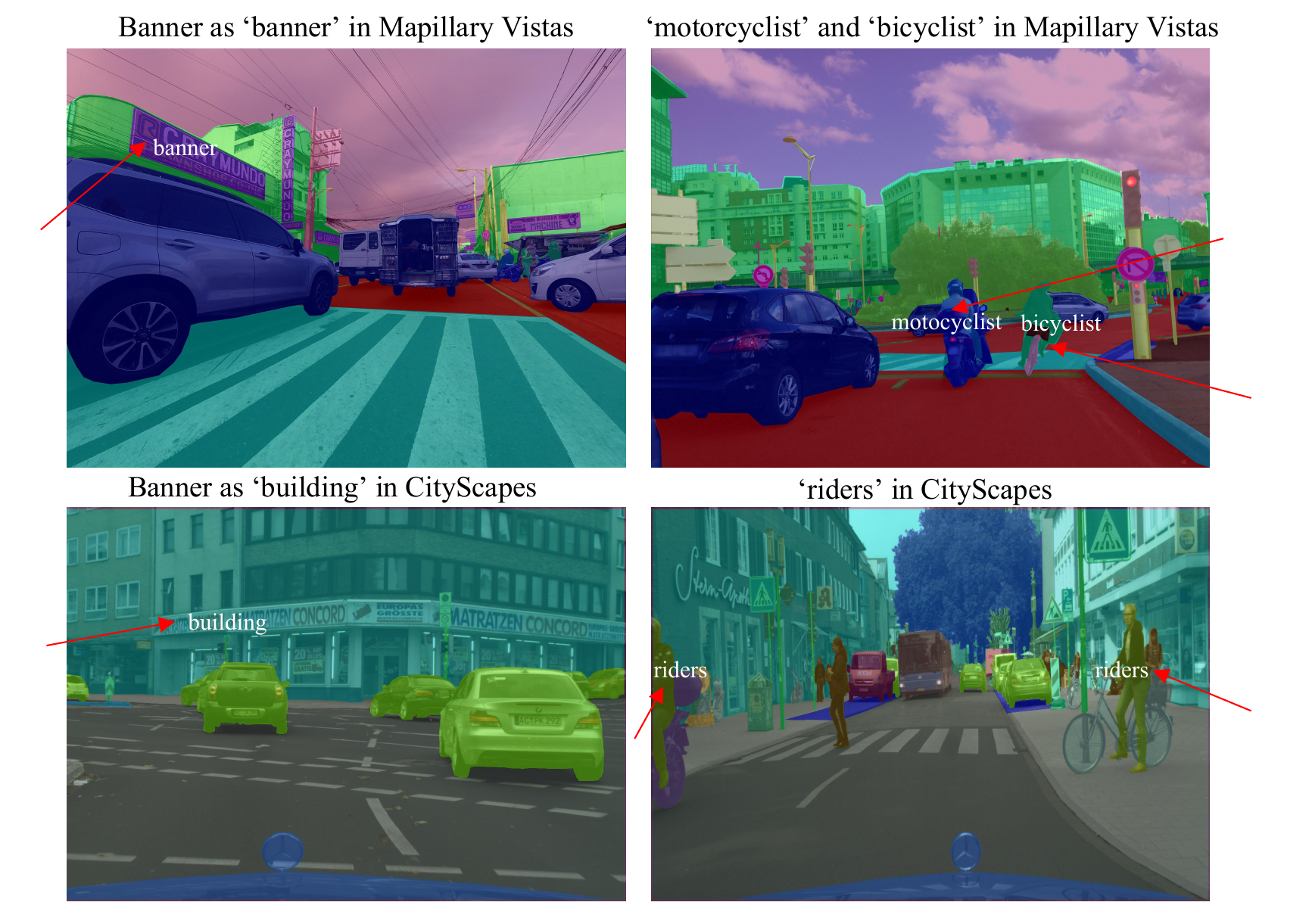}
    \caption{Examples of inconsistent annotations among different datasets. The red arrows point to the instances with inconsistent annotations.}
    \label{fig:my_label}
\end{figure}

\subsection{Taxonomy}\label{data mapping}

%
To train a robust semantic segmentation model that performs well on a variety of datasets, we need a unified taxonomy. Since the above six datasets are obviously inconsistent in category definitions and annotation standards, a naive merge would yield poor segmentation performance~\cite{lambert2020mseg}. Therefore, we follow a sequence of decision rules \cite{bevandic2020multi} to decide on merge and split operations on the taxonomies of the six datasets. Suppose we map all the six component datasets to a unified taxonomy $\chi$. 

\begin{itemize}
\item If a category is defined exactly the same among datasets, $c_i = c_j$, they are merged naively, $\chi = \chi \cup c_i$.
\item If a category is a subset of another category, $c_i \subset c_j$, the superset is replaced by its difference with the subset: $c_d = c_j \setminus c_i, \chi = \chi \cup \{c_d, c_i \}$.
\item If the definitions of the two categories overlap, $c_u = c_i \cap c_j, c_{di} = c_i - c_u \neq \emptyset, c_{dj} = c_j - c_u \neq \emptyset$, then $\chi = \chi \cup c_u \cup c_{di} \cup c_{dj}$.\\
\end{itemize}

Following the above rules, we map all categories in the six datasets to the unified category space, resulting in a total of 194 categories in the merged dataset.

\subsection{Challenge}\label{data problems}

Despite relying on the merged dataset in a unified category space, training a high-performance model that adapts to different domains is still a challenge.
To further understand and fully utilize the merged dataset, we summarize its main characteristics and discuss its existing problems.

\begin{table}[h]
\renewcommand\arraystretch{2.0} 
\begin{center}
\tiny
\setlength{\tabcolsep}{0.25mm}{
\begin{tabular}{lcccc}
\hline \bf Dataset & \bf $\#$Classes & \bf $\#$Images  & \bf Resolution & \bf Scene \\ 
\hline
ADE20K & 150 & 22,210 & \makecell[c]{varying spatial resolution \\ with median 307,200 pixels} & \makecell[c]{Everyday \\ objects}\\
CityScapes & 28 & 3,475 &  $1920 \times 1080$ & \makecell[c]{Driving \\ (Germany)}  \\
MVD & 65 & 20,000 & \makecell[c]{varying spatial resolution \\ at least FullHD} & \makecell[c]{Driving \\(Worldwide)}\\
Scannet & 40 & 24,902 & $1296 \times 968$ & Indoor \\
VIPER & 32 & 18,326 & $1920 \times 1080$ & \makecell[c]{Driving \\(Virtualword)}\\
WildDash & 26 & 4,265 &  $1920 \times 1080$ & \makecell[c]{Driving \\(Worldwide)}\\
\hline
\end{tabular}
}
\end{center}
\caption{\label{data_info} Main characteristics summary of the six datasets in RVC.}
\end{table}

First, one challenge is the data distribution divergence among different domains.
We summarize the main characteristics of the six datasets, which are listed in Table~\ref{data_info}. It shows that there exist significant differences among the six datasets, which are mainly reflected in the imbalance of scene, category, image resolution and dataset size. To further understand the characteristic of the merged dataset that is created in Section~\ref{data mapping}, we analyze the category distribution both in each component dataset and the merged dataset under the unified taxonomy space, which is illustrated in Fig.~\ref{cls_num}. The histogram in Fig.~\ref{cls_num} (a) exhibits that category imbalance of variety and quantity exists in both intra and inter-datasets. Besides, there is a long-tailed distribution of categories in the merged dataset (see Fig.~\ref{cls_num} (b)).

An underlying issue of the merged dataset is the different definitions of the taxonomy and annotation standards in different datasets. 
For example, ‘riders’ in CityScapes is split into ‘motorcyclist’, ‘bicyclist’ and ‘rider\_other’ in Mapillary Vistas.
For another example, ‘banner’ on building is labeled as ‘building’ in CityScapes, but ‘banner' in Mapillary Vistas. Fig.~\ref{fig:my_label} shows some examples of the inconsistent annotations among different datasets. Due to a lack of compatible category supervision, the annotation conflicts and inconsistent category granularity from different datasets substantially reduce the accuracy of models trained across domains. Nevertheless, it is expensive to relabel the conflict categories.

\begin{figure*}[htbp]
\centering
\includegraphics[height=10.4cm, width=16cm]{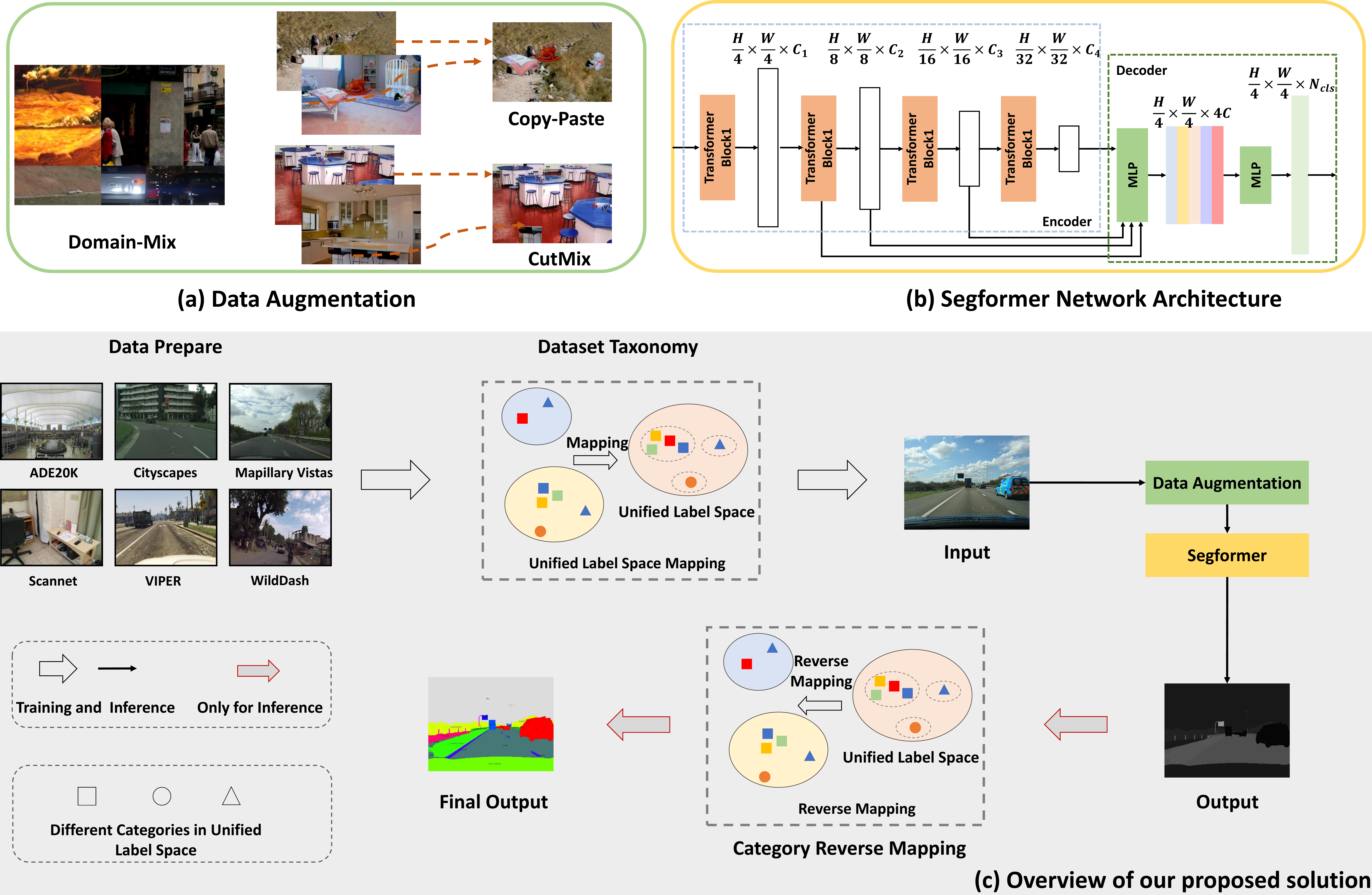}
\caption{An overflow pipeline of our proposed robust semantic segmentation approach for multi-domain learning. (a) Data augmentation details: Copy-Paste, DomainMix and CutMix. (b) The detailed network architecture of Segformer~\cite{DBLP:conf/nips/XieWYAAL21}. (c) The overall pipeline of our proposed solution. }\label{overflow}
\end{figure*}

\section{Methods}\label{methods}

The overall architecture of our method is depicted in Fig.~\ref{overflow}. Given the six benchmarks, we first apply a unified taxonomy on the datasets for cross-domain model training. To enhance the performance of rare categories, we introduce three heuristic data augmentation strategies to increase their frequency of occurrence in training. 
We employ Segformer~\cite{DBLP:conf/nips/XieWYAAL21} as our model architecture. In addition, more designed training strategies are applied to further improve the model generalization. For the output of each benchmark, we obtain the final predictions by summing softmax probabilities of all universal classes that map to the particular dataset class.

\subsection{Data Augmentation}\label{data augmentation}
The categories of the given six semantic segmentation benchmarks and the merged dataset tend to be unbalanced. To improve the performance of rare categories and make the model learn class unbiased, we introduce three heuristic data augmentation approaches (see Fig.~\ref{overflow} (b) for illustration), which are Copy-Paste~\cite{DBLP:journals/corr/abs-2110-05474,DBLP:journals/corr/abs-2012-07177}, DomainMix~\cite{DBLP:journals/corr/abs-2107-13389} and  CutMix~\cite{DBLP:journals/corr/abs-1905-04899}.
In addition, these strategies can increase the diversity of data and prevent overfitting in training.

\noindent\textbf{Copy-Paste.} Copy-Paste~\cite{DBLP:journals/corr/abs-2110-05474,DBLP:journals/corr/abs-2012-07177} constructs efficient data that can improve the poor performance of rare object categories in semantic segmentation. The way it generates a new image is simple and time-friendly: pasting new objects from the source image onto the target image. Copy-Paste computes the binary mask of pasted objects using ground-truth annotations and composes the new images $\hat{I}$ as 
\begin{equation}
\hat{I} = I_1 \times M + I_2 \times (1-M)\label{eq1}
\end{equation}
where $I_1$ is the source image, $I_2$ is the target image and $M$ is the composition mask gained from segmentation annotations.

In our experiments, we first analyze the trained results without data augmentation and select the under-performed (i.e., low accuracy) categories.
The images from under-performed categories are selected as the source images for Copy-Paste augmentation for further training.

\noindent\textbf{DomainMix.} Here we introduce DomainMix~\cite{DBLP:journals/corr/abs-2107-13389} to learn robust feature representations to mitigate domain shifts. DomainMix mixes images from different domains to produce a new image. For a training image, we randomly sample three additional images from the image subset which contains under-performed categories. Then a new mixed image is composed by mixing random crops of these four images into a $2\times2$ gird. DomainMix is a data-efficient strategy that can overcome domain shift and handle category imbalance.

\noindent\textbf{CutMix.} We also adopt CutMix~\cite{DBLP:journals/corr/abs-1905-04899} data augmentation to give a higher sampling probability for rare categories. CutMix strategy aims to generate a new augmented image by replacing the removed regions with a patch from another image. The data augmentation operation in our experiments is defined as
\begin{equation}
\hat{I} = {\rm CutMix}({\rm Crop}(I_2), I_1)\label{eq2}
\end{equation}
where $I_1$ is a source image sampled from under-performed image subset, $I_2$ is a target image, $\hat{I}$ is the augmented image and ${\rm Crop}(\bullet)$ denotes the crop operation. The ${\rm Crop}(\bullet)$ operation is implemented on the patch of the source image $I_2$, which includes the selected under-performed categories.

\subsection{Segmentation Model}\label{Model}
In this section, we introduce our robustness segmentation model learning for characterizing multi-domain data distribution from model architecture, training strategies and category reverse mapping.

\subsubsection{Segformer}
The merged dataset contains six benchmarks with different characteristics, e.g., indoors vs. outdoors, real vs. synthetic, sunny vs. bad weather. So it is necessary to adopt a network with strong representation power. As shown in Fig.~\ref{overflow} (c), Segformer~\cite{DBLP:conf/nips/XieWYAAL21}, an excellent Transformer-based model, is composed of an efficient hierarchical vision Transformer as the encoder and a lightweight multi-layer perception as the decoder. Segformer achieves excellent performance both on ADE20K and Cityscapes. With extensive experiments, we adopt Segformer-MiT-B5 as our segmentation framework for the 2022 RVC semantic segmentation contest.

\subsubsection{Training Strategies}\label{Tricks}

In order to stabilize the training process and improve model accuracy, we adopt some training strategies that are beneficial to multi-domain learning.

\textbf{Class-balanced Sampling}. The merged datasets are extremely imbalanced concerning the number of instances for each class, which forms a long-tailed distribution (as illustrated in Fig.~\ref{cls_num} (b)). The learned segmentation model tends to perform better on instance-rich classes while performing significantly worse on instance-scarce (or tail) classes. To address this issue, we increase the proportion of training samples of the instance-scarce classes at the image level.

\textbf{Warmup}. Due to the inconsistent distribution of the datasets mapped to the unified taxonomy, we use a linear warm-up learning rate schedule to reduce volatility in the early stages of the training~\cite{2017Accurate}. The gradual warm-up strategy contributes to ensuring the stability of the model. The learning rate $lr$ using the warm-up strategy is defined as
\begin{center}
\begin{equation}
\begin{split}
lr = start\_lr  + (end\_lr - start\_lr)\times \frac{cur\_step}{warmup\_steps} \\
\end{split}
\end{equation}
\end{center}
where $start\_lr$ is the initial learning rate, $end\_lr$ is the final learning rate in the warmup stage, $cur\_step$ is the current training step, and $warmup\_steps$ is the total training steps which apply warm-up strategy.

\subsubsection{Reverse Mapping}

\textbf{Many-to-One}. Due to the different category granularity among datasets, some classes in datasets are split into several similar classes in the unified taxonomy, which preserves as many classes as possible. For each class of the raw dataset space, we map all the relevant categories in the unified taxonomy to it when inferring.  For example, when we calculate the final probability of ‘rider’ in CityScapes, we map ‘bicyclist’, ‘motorcyclist’ and ‘rider\_other’ in the unified taxonomy to it. The mapping process can be formulated as
\begin{equation}
    prob_c =\frac{1}{N} \sum\limits_{i\in\phi}^{N}{{\rm Softmax}(\chi_{c_i})}
\end{equation}
where $prob_c$ is the final probability of class $c$ in raw dataset space, $\chi_{c_i}$ is the model output of class $c_i$ in the unified taxonomy space, $\phi$ is the relevant category subset in the unified taxonomy space of class $c$, and $N$ is the number of categories in the subset.

\section{Experiments}\label{exp}
\subsection{Implementation Details}\label{subsec1}
We use Segformer-MiT-B5 as our segmentation model. The encoder is pretrained on the Imagenet-1K dataset and the decoder is randomly initialized. 
We train the model using AdamW optimizer. We spend one epoch linearly increasing the learning rate from 0 to $6\times10^{-5}$. Then, the learning rate is decayed to 0 using a “poly” LR schedule with factor 1.0.

When forming the train data list for an epoch, we evenly sample 10,000 images on the six datasets. If the number of images in datasets is less than 10,000, sampling with replacement is performed. Then we randomly shuffle the generated data, and each minibatch draws 16 samples from it.
During training, we apply data augmentation through random scale jittering with a ratio from 0.5 to 2.0, random horizontal flipping and random cropping to ${960} \times ${640} for all six datasets. 

Our final model is trained on 4 Tesla V100 GPUs. The batch size is set to 16 and the training involves 50 epochs, which takes around 2.5 days. We evaluate our model with the mean Intersection over Union (mIoU) metric.

All test images are evaluated on multiple scales along with image flip process. At test time, we resize the image to five scales with ratio 0.75 to 1.25, implement inference and then interpolate the prediction maps back to the original resolution.

\subsection{Ablation Study}\label{ablation study}
To further understand the impact of data augmentation and training strategies, we perform ablation studies to analyze the contributions of different components. We use Segformer-MiT-B5 architecture as the base model for ablation.
All experiments are evaluated on the validation datasets of the six benchmarks.

\begin{table*}[h]
\begin{center}
\begin{minipage}{750 pt}
\scriptsize
 \setlength{\tabcolsep}{2.2mm}{
\begin{tabular}{ccc|cccccccc}
\hline \bf CP & \bf CM & \bf DM &\bf ADE20K& \bf CityScapes & \bf MVD &\bf Scannet &\bf VIPER &\bf  WildDash &\textit h.mean \\ 
\hline
\qquad  &\qquad & \qquad  & 46.62 & 76.38 & 50.13 & 79.66 & 66.46 & 65.15 & 64.07  \\
\checkmark & \qquad & \qquad & 46.64 & 76.31 & 50.57 & 79.26 & 67.52 & 64.82 & 64.19 \\
\qquad & \checkmark & \qquad & 46.90 & 76.66 & 51.12 & 79.55 & 67.43 & 65.55 & 64.54 \\
\qquad & \qquad & \checkmark & 46.92 & 76.52 & 50.72 & \bf79.96 & 67.30 & 65.64 & 64.51 \\
\checkmark & \qquad & \checkmark & \bf47.19 & 76.77 & \bf51.21 & 79.67 & \bf68.24 & \bf65.82 &  \bf64.82\\
\qquad & \checkmark & \checkmark & 47.12 & \bf76.86 & 50.56 & 79.25 & 67.72 & 65.56 &  64.51\\
\checkmark & \checkmark & \checkmark & 46.89 & 76.74 & 50.40 & 79.89 & 67.26 & 65.81 & 64.50 \\
\hline
\end{tabular}}
\end{minipage}
\end{center}
\caption{\label{da-ablation} Quantitative ablations on the contributions of different data augmentation combinations: Copy-Paste(CP), CutMix(CM), DomainMix(DM).}
\end{table*}

\noindent\textbf{Data augmentation.} We evaluate the contributions of three data augmentation strategies by removing them one by one. To understand the comprehensive performance of the data augmentation strategies, we further add the harmonic mean mIoU (\textit h.mean) computed by six datasets. 

The quantitative results of different data augmentation combinations are displayed in Table~\ref{da-ablation}. Compared with the performance of the baseline that was evaluated by \textit h.mean, all data augmentation strategies achieve model performance gains. It means that the under-performed categories are given more chances to be sampled using these data augmentation strategies, which can alleviate the category imbalance and the biased training issue of the merged dataset.
Besides, the Domain-Mix data augmentation, which mixes images from different domains into a new image, can overcome domain shift without using special network structures.
We apply the data augmentation combination of Copy-Paste and DomainMix in our final RVC 2022 submission, which obtains the best performance in Table~\ref{da-ablation}.
\\

\noindent\textbf{Training strategies.} We perform ablations on different training strategies to verify their contributions. Table~\ref{Trick-ablation} shows the experiment results. The baseline is a Segformer-MiT-B5 trained without any training strategy. Then we add the training strategies introduced in Section 4.2 to the baseline. As shown, class-balanced sampling and warmup learning rate schedule added to the baseline respectively both lead to performance gains. And combining these two strategies boosts the performance of the baseline by nearly 1\% in mIoU.

\begin{table*}[h]
\begin{center}
\begin{minipage}{470pt}
\scriptsize
\setlength{\tabcolsep}{1.5mm}{
\begin{tabular}{cc|cccccccc}
\hline \bf Class-balanced & \bf Warmup  &\bf ADE20K& \bf CityScapes & \bf MVD &\bf Scannet &\bf VIPER &\bf  WildDash &\textit h.mean \\ 
\hline
\qquad & \qquad & 46.0 & 76.4 & 50.3 & 79.8 & 66.6 & 65.8 & 64.1 \\
\checkmark & \qquad & 46.2 & 76.7 & 50.4 & \bf80.0 & 66.6 & 65.5 & 64.2 \\
\qquad & \checkmark & 46.5 & \bf76.9 & 49.9 & 79.8 & 66.4 & 65.7 & 64.2 \\
\checkmark & \checkmark& \bf47.2 & 76.8 & \bf51.2 & 79.7 & \bf68.2 & \bf65.8 & \bf64.8 \\
\hline
\end{tabular}}
\end{minipage}
\end{center}
\caption{\label{Trick-ablation} Quantitative ablations on the contributions of different training strategies.}
\end{table*}

\subsection{Model Selection for Robust Learning}\label{model selection}

We consider that high model capacity can reduce the impact of inconsistent data distribution on the performance of the multi-domain model. For example, compared with DeepLabv3+/Resnet50, DeepLabv3+/Resnet18 with smaller capacity would lead to a more significant performance decrease when datasets with different distributions are introduced to the training stage. Put another way, it is difficult for a model with insufficient capacity to adapt to datasets with different characteristics. As a result, it would lead to a decrease in the accuracy of each dataset compared to when training alone. 
To further understand our hypothesis, we design an incremental experiment on the six benchmarks in the unified taxonomy space.

For a fair comparison, we select Segformer-MiT-B1 and DeepLabv3+/Resnet50 to conduct the incremental experiment, as they achieve similar performance on the ADE20K dataset (i.e., 42.72 mIoU vs. 42.13 mIoU). 
We also choose a larger capacity model Segformer-MiT-B5 that achieves better performance on ADE20K dataset (50.19 mIoU) for further comparison. The experiment is implemented by adding the datasets which map to the unified label space into the semantic segmentation training  progressively. 

The incremental experimental results are shown in Table~\ref{model-cp}. When we gradually add Cityscapes, WildDash, MVD and VIPER to the training process, the prediction accuracy of all the three model architectures in each dataset is improved slightly, which is probably thanks to the similar scenes and categories in these datasets.
Moreover, when ADE20k and Scannet are added, the accuracy of the three model architectures all decrease in the initial four training datasets such as CityScapes. It may be due to the increase of categories in training and the large divergence of these datasets.

However, compared with Segformer-MiT-B1 and DeepLabv3+/Resnet50 with more accuracy drop, Segformer-MiT-B5 with larger model capacity is insensitive to data distribution inconsistence and has better generalization ability across domains.
Therefore, we select a Transformer-based model with high representation ability for multi-domain semantic segmentation learning.

\begin{table*}
\setlength\aboverulesep{0pt}
\scriptsize
\begin{center}
\resizebox{\textwidth}{!}{
\begin{tabular}{l|cccccc}
\hline \bf Model & \bf CityScapes & \bf WildDash & \bf MVD &\bf VIPER &\bf ADE20K &\bf Scannet \\ 
\hline
\multirow{6}*{DeepLabv3+/Resnet50} & 70.0 & - & - & - & - & - \\
~ & 71.3 & 59.5 & - & - & - & - \\
~ & 73.8 & 61.7 & 44.1 & - & - & - \\
~ & 74.0 & 61.9 & 44.7 & 61.6 & - & - \\
~ & 72.6 & 60.8 & 43.2 & 59.8 & 33.3 & - \\
~ & 71.1 & 58.4 & 37.2 & 53.6 & 28.6 & 65.2 \\
\hline
\multirow{6}*{Segformer-MiT-B1} & 72.0 & - & - & - & - & - \\
~ & 71.3 & 59.0 & - & - & - & - \\
~ & 72.7 & 59.6 & 45.4 & - & - & - \\
~ & 72.6 & 60.2 & 45.7 & 62.4 & - & - \\
~ & 70.7 & 59.4 & 43.8 & 60.3 & 37.4 & - \\
~ & 70.4 & 58.8 & 41.3 & 59.7 & 36.6 & 72.8 \\
\hline
\multirow{6}*{Segformer-MiT-B5} & 76.9 & - & - & - & - & - \\
~ & 76.9 & 65.2 & - & - & - & - \\
~ & 77.8 & 66.9 & 51.9 & - & - & - \\
~ & 77.8 & 66.9 & 51.6 & 68.9 & - & - \\
~ & 77.1 & 65.3 & 51.0 & 67.9 & 47.3 & - \\
~ & 76.4 & 65.1 & 50.1 & 66.5 & 46.6 & 79.6 \\
\hline
\end{tabular}
}
\end{center}
\caption{\label{model-cp} Results on accuracy (mIoU) of DeepLabv3+/Resnet50, Segformer-MiT-B1, Segformer-MiT-B5 in dataset incremental experiment. CityScapes, WildDash, MVD, VIPER, ADE20K, Scannet are added to the training process one by one. ‘-’ means that the corresponding dataset does not participate in training.}
\end{table*}

\subsection{Comparison with Domain Generalization Methods}\label{comparsion dg}
Domain generalization (DG) algorithms learn a unified model from several different but related domains. For multi-domain robust semantic segmentation task, we apply some state-of-the-art domain generalization methods to explore the impact of feature alignment on model performance across different benchmarks. In order to comprehensively analyze the impact of domain generalization methods on model performance, we conduct experiments based on some representative methods, which are DANN~\cite{DBLP:journals/corr/GaninUAGLLML15}, pOSAL~\cite{DBLP:journals/tmi/WangYYFH19}, MMD~\cite{DBLP:journals/corr/abs-1909-08531}, IRM~\cite{https://doi.org/10.48550/arxiv.1907.02893} and IIB~\cite{DBLP:conf/aaai/LiSWZRLK022}.

For a quick verification of these algorithms, we select DeepLabv3+/Resnet50 as our model architecture. 
Table~\ref{dg-compare} lists the comparison results with state-of-the-art domain generalization methods on the validation dataset of six benchmarks.
The results indicate that model performance is inconsistent across different benchmarks.
We also summarize the overall performance by the harmonic mean across the six datasets.
However, the harmonic mean mIoU shows all domain generation methods we tried fail to improve the model performance substantially. We presume that the alignment of features among different domains may damage the semantic information and feature discrimination of the segmentation task.

\begin{table*}[h]
\begin{center}
\scriptsize
\begin{minipage}{470pt}
 \setlength{\tabcolsep}{3.0mm}{
\begin{tabular}{l|ccccccc}
\hline \bf Method & \bf ADE20K& \bf CityScapes & \bf MVD &\bf Scannet &\bf VIPER &\bf  WildDash &\textit h.mean\\ 
\hline
Baseline & 26.4 & 66.7 & 35.5 & 57.7 & 53.4 & 57.9 & 49.6\\
DANN ~\cite{DBLP:journals/corr/GaninUAGLLML15} & 24.2 & 64.2 & 33.9 & 56.7 & 52.5 & 57.0 & 48.1 \\
pOSAL~\cite{DBLP:journals/tmi/WangYYFH19} & 25.5 & 64.6 & 33.7 & 57.0 & 52.3 & 55.1 & 48.0\\
MMD ~\cite{DBLP:journals/corr/abs-1909-08531} & 26.3 & 65.1 & 34.4 & 58.1 & 52.4 & 56.3 & 48.8\\
IRM ~\cite{https://doi.org/10.48550/arxiv.1907.02893} & 22.0 & 47.4 & 20.2 & 51.9 & 47.0 & 37.6 & 37.7\\
IIB ~\cite{DBLP:conf/aaai/LiSWZRLK022}& 23.4 & 61.7 & 34.6 & 52.7 & 55.9 & 52.9 & 48.9\\
\hline
\end{tabular}}
\end{minipage}
\end{center}
\caption{\label{dg-compare} Semantic segmentation accuracy (mIoU) of state-of-the-art domain generalization methods.}
\end{table*}

\subsection{Semantic Segmentation Contest of RVC 2022 }\label{rvc2022}
\begin{figure*}
    \centering
    \includegraphics[height=8.26cm, width=16cm]{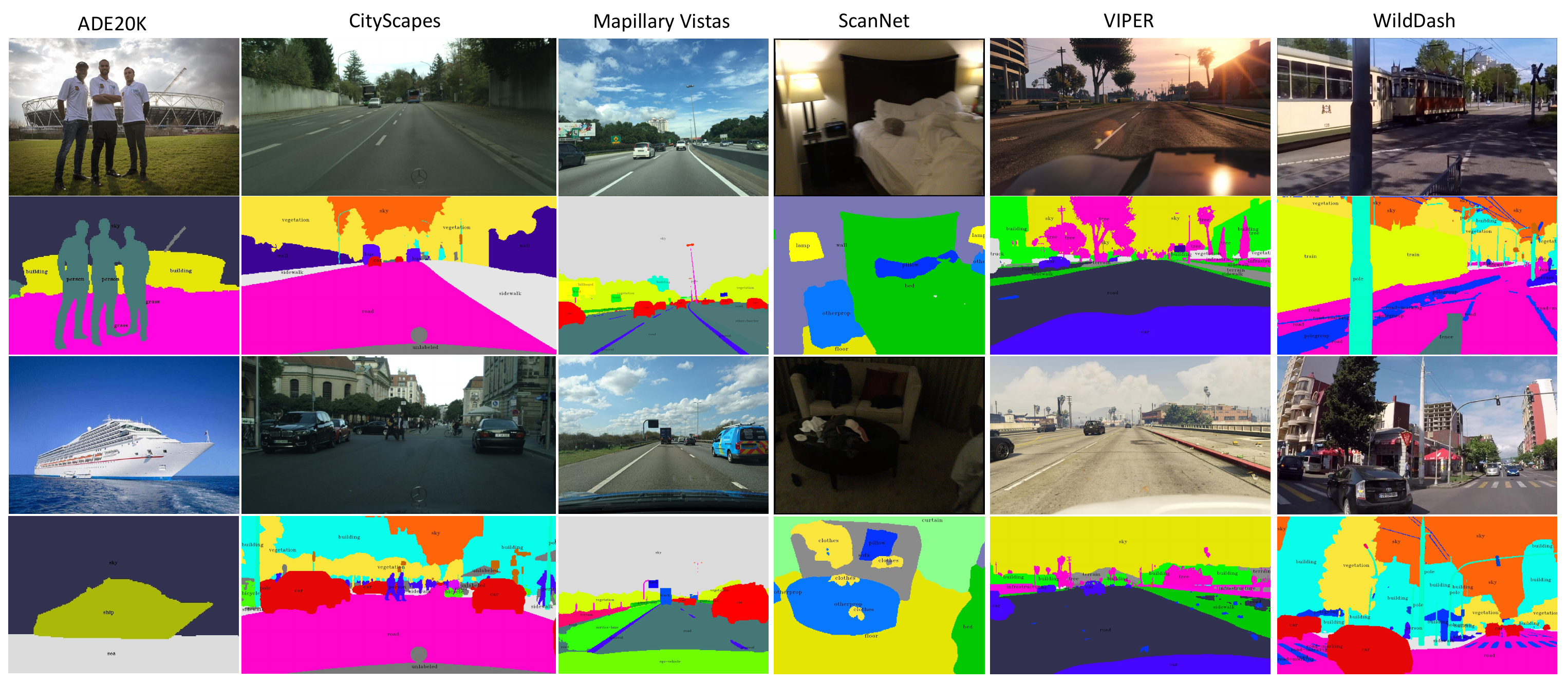}
    \caption{The qualitative semantic segmentation results on images of different datasets.}
    \label{fig:result}
\end{figure*}

Our study starts with the Robust Vision Challenge held at ECCV 2022. RVC 2022 proposes a contest on semantic segmentation task, which aims to train a single model that performs well on multiple datasets with different characteristics. The qualitative semantic segmentation results of our solution in different datasets are shown in Fig.~\ref{fig:result}.

The top three results of participants are represented in Table~\ref{submission}, where first place outperforms our submission except for the WildDash benchmark. However, there is a large discrepancy in hardware and dataset scale between ours and the FAN\_NV\_RVC method. Our final model is trained on 4 Tesla V100 GPUs with batch size 16, yet FAN\_NV\_RVC applies 64 Tesla V100 GPUs with batch size 64. The resolution in the training of FAN\_NV\_RVC, $1024\times1024$, is also larger than our resolution $960\times640$. Moreover, FAN\_NV\_RVC adopts additional datasets, including COCO~\cite{DBLP:conf/eccv/LinMBHPRDZ14}, IDD~\cite{DBLP:conf/wacv/VarmaSNCJ19} and BDD~\cite{DBLP:journals/corr/abs-1805-04687}. Compared to our method, FAN\_NV\_RVC utilizes a larger scale of data for training, i.e., 84,108 \emph{vs} 223,369. In contrast, our solution can achieve considerable performance at a small cost.

Moreover, the WildDash benchmark includes some unusual and hard scenarios of road scenes, which specifically evaluates the robustness of semantic segmentation models. 
Our approach ranks 1st in the WildDash leaderboard, which shows our model can handle the out-of-distribution problem with robust generalization ability.

\begin{table*}[h]
\begin{center}
\scriptsize
 \setlength{\tabcolsep}{3.0mm}{
\begin{tabular}{lccc}
\hline \bf Dataset & \bf FAN\_NV\_RVC& \bf MIX6D\_RVC(Ours) & \bf UDSSEG\_RVC\\ 
\hline
ADE20K & \bf43.5 & 39.1 & 31.6 \\
CityScapes & \bf82.0 & 79.8 & 79.4  \\
MVD & \bf55.3 & 50.9 & 47.4 \\
Scannet & \bf58.6 & 58.2 & 54.5 \\
VIPER & \bf69.8 & 66.5 & 65.2\\
WildDash & 47.5 & \bf48.5 & 45.5 \\
\hline
\end{tabular}}
\end{center}
\caption{\label{submission} Performance (mIoU) of the top three (first: FAN\_NV\_RVC, second: MIX6D\_RVC, third: UDSSEG\_RVC) submissions of RVC 2022 semantic segmentation contest. The competition results are available at \href{http://robustvision.net/leaderboard.php?benchmark=semantic}}
\end{table*}

\section{Conclusions}\label{sec5}
This paper presents an empirical study on robust multi-domain learning of semantic segmentation task across datasets with different characteristics.
We analyzed the main characteristics of our merged dataset and discuss its existing challenges.
Furthermore, we concluded the learning techniques that are beneficial for model generalization, including data augmentation, training strategies, and model capacity.
These discoveries help our model, which is trained at a small cost, rank 2nd in the semantic segmentation task of ECCV Robust Vision Challenge 2022.
Moreover, we explored the impact of some domain generalization algorithms on multi-domain learning but failed to bring performance gains. Extensive experiments on DG approaches demonstrate that there is probably a gap between the two tasks. 
How to effectively use multiple datasets with different characteristics to train a robust model is of great significance. Therefore, we hope our work can provide a baseline and lead to further research in this realm.


\bibliography{sn-bibliography}


\begin{thebibliography}{43}
\ifx \bisbn   \undefined \def \bisbn  #1{ISBN #1}\fi
\ifx \binits  \undefined \def \binits#1{#1}\fi
\ifx \bauthor  \undefined \def \bauthor#1{#1}\fi
\ifx \batitle  \undefined \def \batitle#1{#1}\fi
\ifx \bjtitle  \undefined \def \bjtitle#1{#1}\fi
\ifx \bvolume  \undefined \def \bvolume#1{\textbf{#1}}\fi
\ifx \byear  \undefined \def \byear#1{#1}\fi
\ifx \bissue  \undefined \def \bissue#1{#1}\fi
\ifx \bfpage  \undefined \def \bfpage#1{#1}\fi
\ifx \blpage  \undefined \def \blpage #1{#1}\fi
\ifx \burl  \undefined \def \burl#1{\textsf{#1}}\fi
\ifx \doiurl  \undefined \def \doiurl#1{\url{https://doi.org/#1}}\fi
\ifx \betal  \undefined \def \betal{\textit{et al.}}\fi
\ifx \binstitute  \undefined \def \binstitute#1{#1}\fi
\ifx \binstitutionaled  \undefined \def \binstitutionaled#1{#1}\fi
\ifx \bctitle  \undefined \def \bctitle#1{#1}\fi
\ifx \beditor  \undefined \def \beditor#1{#1}\fi
\ifx \bpublisher  \undefined \def \bpublisher#1{#1}\fi
\ifx \bbtitle  \undefined \def \bbtitle#1{#1}\fi
\ifx \bedition  \undefined \def \bedition#1{#1}\fi
\ifx \bseriesno  \undefined \def \bseriesno#1{#1}\fi
\ifx \blocation  \undefined \def \blocation#1{#1}\fi
\ifx \bsertitle  \undefined \def \bsertitle#1{#1}\fi
\ifx \bsnm \undefined \def \bsnm#1{#1}\fi
\ifx \bsuffix \undefined \def \bsuffix#1{#1}\fi
\ifx \bparticle \undefined \def \bparticle#1{#1}\fi
\ifx \barticle \undefined \def \barticle#1{#1}\fi
\bibcommenthead
\ifx \bconfdate \undefined \def \bconfdate #1{#1}\fi
\ifx \botherref \undefined \def \botherref #1{#1}\fi
\ifx \url \undefined \def \url#1{\textsf{#1}}\fi
\ifx \bchapter \undefined \def \bchapter#1{#1}\fi
\ifx \bbook \undefined \def \bbook#1{#1}\fi
\ifx \bcomment \undefined \def \bcomment#1{#1}\fi
\ifx \oauthor \undefined \def \oauthor#1{#1}\fi
\ifx \citeauthoryear \undefined \def \citeauthoryear#1{#1}\fi
\ifx \endbibitem  \undefined \def \endbibitem {}\fi
\ifx \bconflocation  \undefined \def \bconflocation#1{#1}\fi
\ifx \arxivurl  \undefined \def \arxivurl#1{\textsf{#1}}\fi
\csname PreBibitemsHook\endcsname

\bibitem{lambert2020mseg}
\begin{bchapter}
\bauthor{\bsnm{Lambert}, \binits{J.}},
\bauthor{\bsnm{Liu}, \binits{Z.}},
\bauthor{\bsnm{Sener}, \binits{O.}},
\bauthor{\bsnm{Hays}, \binits{J.}},
\bauthor{\bsnm{Koltun}, \binits{V.}}:
\bctitle{Mseg: A composite dataset for multi-domain semantic segmentation}.
In: \bbtitle{Proceedings of the IEEE/CVF Conference on Computer Vision and
  Pattern Recognition},
pp. \bfpage{2879}--\blpage{2888}
(\byear{2020})
\end{bchapter}
\endbibitem

\bibitem{DBLP:journals/corr/GaninUAGLLML15}
\begin{botherref}
\oauthor{\bsnm{Ganin}, \binits{Y.}},
\oauthor{\bsnm{Ustinova}, \binits{E.}},
\oauthor{\bsnm{Ajakan}, \binits{H.}},
\oauthor{\bsnm{Germain}, \binits{P.}},
\oauthor{\bsnm{Larochelle}, \binits{H.}},
\oauthor{\bsnm{Laviolette}, \binits{F.}},
\oauthor{\bsnm{Marchand}, \binits{M.}},
\oauthor{\bsnm{Lempitsky}, \binits{V.S.}}:
Domain-adversarial training of neural networks.
CoRR
\textbf{abs/1505.07818}
(2015)
\end{botherref}
\endbibitem

\bibitem{DBLP:conf/cvpr/IsolaZZE17}
\begin{bchapter}
\bauthor{\bsnm{Isola}, \binits{P.}},
\bauthor{\bsnm{Zhu}, \binits{J.}},
\bauthor{\bsnm{Zhou}, \binits{T.}},
\bauthor{\bsnm{Efros}, \binits{A.A.}}:
\bctitle{Image-to-image translation with conditional adversarial networks}.
In: \bbtitle{2017 {IEEE} Conference on Computer Vision and Pattern Recognition,
  {CVPR} 2017, Honolulu, HI, USA, July 21-26, 2017},
pp. \bfpage{5967}--\blpage{5976}
(\byear{2017})
\end{bchapter}
\endbibitem

\bibitem{DBLP:conf/cvpr/LiPWK18}
\begin{bchapter}
\bauthor{\bsnm{Li}, \binits{H.}},
\bauthor{\bsnm{Pan}, \binits{S.J.}},
\bauthor{\bsnm{Wang}, \binits{S.}},
\bauthor{\bsnm{Kot}, \binits{A.C.}}:
\bctitle{Domain generalization with adversarial feature learning}.
In: \bbtitle{2018 {IEEE} Conference on Computer Vision and Pattern Recognition,
  {CVPR} 2018, Salt Lake City, UT, USA, June 18-22, 2018},
pp. \bfpage{5400}--\blpage{5409}
(\byear{2018})
\end{bchapter}
\endbibitem

\bibitem{DBLP:conf/cvpr/GongLCG19}
\begin{bchapter}
\bauthor{\bsnm{Gong}, \binits{R.}},
\bauthor{\bsnm{Li}, \binits{W.}},
\bauthor{\bsnm{Chen}, \binits{Y.}},
\bauthor{\bsnm{Gool}, \binits{L.V.}}:
\bctitle{{DLOW:} domain flow for adaptation and generalization}.
In: \bbtitle{{IEEE} Conference on Computer Vision and Pattern Recognition,
  {CVPR} 2019, Long Beach, CA, USA, June 16-20, 2019},
pp. \bfpage{2477}--\blpage{2486}
(\byear{2019})
\end{bchapter}
\endbibitem

\bibitem{DBLP:conf/eccv/LiTGLLZT18}
\begin{botherref}
\oauthor{\bsnm{Li}, \binits{Y.}},
\oauthor{\bsnm{Tian}, \binits{X.}},
\oauthor{\bsnm{Gong}, \binits{M.}},
\oauthor{\bsnm{Liu}, \binits{Y.}},
\oauthor{\bsnm{Liu}, \binits{T.}},
\oauthor{\bsnm{Zhang}, \binits{K.}},
\oauthor{\bsnm{Tao}, \binits{D.}}:
Deep domain generalization via conditional invariant adversarial networks.
In: Ferrari, V., Hebert, M., Sminchisescu, C., Weiss, Y. (eds.)
Computer Vision - {ECCV} 2018 - 15th European Conference, Munich, Germany,
  September 8-14, 2018, Proceedings, Part {XV}.
Lecture Notes in Computer Science,
vol. 11219,
pp. 647--663
\end{botherref}
\endbibitem

\bibitem{DBLP:conf/cvpr/ShaoLLY19}
\begin{botherref}
\oauthor{\bsnm{Shao}, \binits{R.}},
\oauthor{\bsnm{Lan}, \binits{X.}},
\oauthor{\bsnm{Li}, \binits{J.}},
\oauthor{\bsnm{Yuen}, \binits{P.C.}}:
Multi-adversarial discriminative deep domain generalization for face
  presentation attack detection.
In: {IEEE} Conference on Computer Vision and Pattern Recognition, {CVPR} 2019,
  Long Beach, CA, USA, June 16-20, 2019,
pp. 10023--10031
\end{botherref}
\endbibitem

\bibitem{DBLP:journals/corr/abs-1909-08531}
\begin{botherref}
\oauthor{\bsnm{Wang}, \binits{J.}},
\oauthor{\bsnm{Chen}, \binits{Y.}},
\oauthor{\bsnm{Feng}, \binits{W.}},
\oauthor{\bsnm{Yu}, \binits{H.}},
\oauthor{\bsnm{Huang}, \binits{M.}},
\oauthor{\bsnm{Yang}, \binits{Q.}}:
Transfer learning with dynamic distribution adaptation.
CoRR
\textbf{abs/1909.08531}
(2019)
\end{botherref}
\endbibitem

\bibitem{DBLP:journals/corr/TzengHZSD14}
\begin{botherref}
\oauthor{\bsnm{Tzeng}, \binits{E.}},
\oauthor{\bsnm{Hoffman}, \binits{J.}},
\oauthor{\bsnm{Zhang}, \binits{N.}},
\oauthor{\bsnm{Saenko}, \binits{K.}},
\oauthor{\bsnm{Darrell}, \binits{T.}}:
Deep domain confusion: Maximizing for domain invariance.
CoRR
\textbf{abs/1412.3474}
(2014)
\end{botherref}
\endbibitem

\bibitem{DBLP:conf/mm/WangFCYHY18}
\begin{botherref}
\oauthor{\bsnm{Wang}, \binits{J.}},
\oauthor{\bsnm{Feng}, \binits{W.}},
\oauthor{\bsnm{Chen}, \binits{Y.}},
\oauthor{\bsnm{Yu}, \binits{H.}},
\oauthor{\bsnm{Huang}, \binits{M.}},
\oauthor{\bsnm{Yu}, \binits{P.S.}}:
Visual domain adaptation with manifold embedded distribution alignment.
In: Boll, S., Lee, K.M., Luo, J., Zhu, W., Byun, H., Chen, C.W., Lienhart, R.,
  Mei, T. (eds.)
2018 {ACM} Multimedia Conference on Multimedia Conference, {MM} 2018, Seoul,
  Republic of Korea, October 22-26, 2018,
pp. 402--410
\end{botherref}
\endbibitem

\bibitem{DBLP:conf/wacv/VarmaSNCJ19}
\begin{botherref}
\oauthor{\bsnm{Varma}, \binits{G.}},
\oauthor{\bsnm{Subramanian}, \binits{A.}},
\oauthor{\bsnm{Namboodiri}, \binits{A.M.}},
\oauthor{\bsnm{Chandraker}, \binits{M.}},
\oauthor{\bsnm{Jawahar}, \binits{C.V.}}:
{IDD:} {A} dataset for exploring problems of autonomous navigation in
  unconstrained environments.
In: {IEEE} Winter Conference on Applications of Computer Vision, {WACV} 2019,
  Waikoloa Village, HI, USA, January 7-11, 2019,
pp. 1743--1751
\end{botherref}
\endbibitem

\bibitem{DBLP:conf/iccv/BerrielLNKOS019}
\begin{botherref}
\oauthor{\bsnm{Berriel}, \binits{R.F.}},
\oauthor{\bsnm{Lathuili{\`{e}}re}, \binits{S.}},
\oauthor{\bsnm{Nabi}, \binits{M.}},
\oauthor{\bsnm{Klein}, \binits{T.}},
\oauthor{\bsnm{Oliveira{-}Santos}, \binits{T.}},
\oauthor{\bsnm{Sebe}, \binits{N.}},
\oauthor{\bsnm{Ricci}, \binits{E.}}:
Budget-aware adapters for multi-domain learning.
In: 2019 {IEEE/CVF} International Conference on Computer Vision, {ICCV} 2019,
  Seoul, Korea (South), October 27 - November 2, 2019,
pp. 382--391
\end{botherref}
\endbibitem

\bibitem{DBLP:conf/cvpr/XiaoG020}
\begin{botherref}
\oauthor{\bsnm{Xiao}, \binits{J.}},
\oauthor{\bsnm{Gu}, \binits{S.}},
\oauthor{\bsnm{Zhang}, \binits{L.}}:
Multi-domain learning for accurate and few-shot color constancy.
In: 2020 {IEEE/CVF} Conference on Computer Vision and Pattern Recognition,
  {CVPR} 2020, Seattle, WA, USA, June 13-19, 2020,
pp. 3255--3264
\end{botherref}
\endbibitem

\bibitem{DBLP:conf/nips/RebuffiBV17}
\begin{bchapter}
\bauthor{\bsnm{Rebuffi}, \binits{S.}},
\bauthor{\bsnm{Bilen}, \binits{H.}},
\bauthor{\bsnm{Vedaldi}, \binits{A.}}:
\bctitle{Learning multiple visual domains with residual adapters}.
In: \beditor{\bsnm{Guyon}, \binits{I.}},
\beditor{\bparticle{von} \bsnm{Luxburg}, \binits{U.}},
\beditor{\bsnm{Bengio}, \binits{S.}},
\beditor{\bsnm{Wallach}, \binits{H.M.}},
\beditor{\bsnm{Fergus}, \binits{R.}},
\beditor{\bsnm{Vishwanathan}, \binits{S.V.N.}},
\beditor{\bsnm{Garnett}, \binits{R.}} (eds.)
\bbtitle{Advances in Neural Information Processing Systems 30: Annual
  Conference on Neural Information Processing Systems 2017, December 4-9, 2017,
  Long Beach, CA, {USA}},
pp. \bfpage{506}--\blpage{516}
(\byear{2017})
\end{bchapter}
\endbibitem

\bibitem{DBLP:journals/pami/RosenfeldT20}
\begin{barticle}
\bauthor{\bsnm{Rosenfeld}, \binits{A.}},
\bauthor{\bsnm{Tsotsos}, \binits{J.K.}}:
\batitle{Incremental learning through deep adaptation}.
\bjtitle{{IEEE} Trans. Pattern Anal. Mach. Intell.}
\bvolume{42}(\bissue{3}),
\bfpage{651}--\blpage{663}
(\byear{2020})
\end{barticle}
\endbibitem

\bibitem{DBLP:journals/corr/BilenV17}
\begin{botherref}
\oauthor{\bsnm{Bilen}, \binits{H.}},
\oauthor{\bsnm{Vedaldi}, \binits{A.}}:
Universal representations: The missing link between faces, text, planktons, and
  cat breeds.
CoRR
\textbf{abs/1701.07275}
(2017)
\end{botherref}
\endbibitem

\bibitem{DBLP:conf/nips/XieWYAAL21}
\begin{botherref}
\oauthor{\bsnm{Xie}, \binits{E.}},
\oauthor{\bsnm{Wang}, \binits{W.}},
\oauthor{\bsnm{Yu}, \binits{Z.}},
\oauthor{\bsnm{Anandkumar}, \binits{A.}},
\oauthor{\bsnm{Alvarez}, \binits{J.M.}},
\oauthor{\bsnm{Luo}, \binits{P.}}:
Segformer: Simple and efficient design for semantic segmentation with
  transformers.
In: Ranzato, M., Beygelzimer, A., Dauphin, Y.N., Liang, P., Vaughan, J.W.
  (eds.)
Advances in Neural Information Processing Systems 34: Annual Conference on
  Neural Information Processing Systems 2021, NeurIPS 2021, December 6-14,
  2021, Virtual,
pp. 12077--12090
\end{botherref}
\endbibitem

\bibitem{DBLP:conf/cvpr/HeZRS16}
\begin{botherref}
\oauthor{\bsnm{He}, \binits{K.}},
\oauthor{\bsnm{Zhang}, \binits{X.}},
\oauthor{\bsnm{Ren}, \binits{S.}},
\oauthor{\bsnm{Sun}, \binits{J.}}:
Deep residual learning for image recognition.
In: 2016 {IEEE} Conference on Computer Vision and Pattern Recognition, {CVPR}
  2016, Las Vegas, NV, USA, June 27-30, 2016,
pp. 770--778
\end{botherref}
\endbibitem

\bibitem{DBLP:journals/corr/abs-2107-13389}
\begin{botherref}
\oauthor{\bsnm{Ramamonjison}, \binits{R.}},
\oauthor{\bsnm{Banitalebi{-}Dehkordi}, \binits{A.}},
\oauthor{\bsnm{Kang}, \binits{X.}},
\oauthor{\bsnm{Bai}, \binits{X.}},
\oauthor{\bsnm{Zhang}, \binits{Y.}}:
Simrod: {A} simple adaptation method for robust object detection.
CoRR
\textbf{abs/2107.13389}
(2021)
\end{botherref}
\endbibitem

\bibitem{DBLP:journals/corr/abs-1905-04899}
\begin{botherref}
\oauthor{\bsnm{Yun}, \binits{S.}},
\oauthor{\bsnm{Han}, \binits{D.}},
\oauthor{\bsnm{Oh}, \binits{S.J.}},
\oauthor{\bsnm{Chun}, \binits{S.}},
\oauthor{\bsnm{Choe}, \binits{J.}},
\oauthor{\bsnm{Yoo}, \binits{Y.}}:
Cutmix: Regularization strategy to train strong classifiers with localizable
  features.
CoRR
\textbf{abs/1905.04899}
(2019)
\end{botherref}
\endbibitem

\bibitem{9709604}
\begin{botherref}
\oauthor{\bsnm{Rao}, \binits{Z.}},
\oauthor{\bsnm{Dai}, \binits{Y.}},
\oauthor{\bsnm{Shen}, \binits{Z.}},
\oauthor{\bsnm{He}, \binits{R.}}:
Rethinking training strategy in stereo matching.
IEEE Transactions on Neural Networks and Learning Systems,
1--14
(2022)
\end{botherref}
\endbibitem

\bibitem{DBLP:journals/corr/abs-2110-05474}
\begin{botherref}
\oauthor{\bsnm{Hu}, \binits{H.}},
\oauthor{\bsnm{Wei}, \binits{F.}},
\oauthor{\bsnm{Hu}, \binits{H.}},
\oauthor{\bsnm{Ye}, \binits{Q.}},
\oauthor{\bsnm{Cui}, \binits{J.}},
\oauthor{\bsnm{Wang}, \binits{L.}}:
Semi-supervised semantic segmentation via adaptive equalization learning.
CoRR
\textbf{abs/2110.05474}
(2021)
\end{botherref}
\endbibitem

\bibitem{DBLP:journals/corr/abs-2012-07177}
\begin{botherref}
\oauthor{\bsnm{Ghiasi}, \binits{G.}},
\oauthor{\bsnm{Cui}, \binits{Y.}},
\oauthor{\bsnm{Srinivas}, \binits{A.}},
\oauthor{\bsnm{Qian}, \binits{R.}},
\oauthor{\bsnm{Lin}, \binits{T.}},
\oauthor{\bsnm{Cubuk}, \binits{E.D.}},
\oauthor{\bsnm{Le}, \binits{Q.V.}},
\oauthor{\bsnm{Zoph}, \binits{B.}}:
Simple copy-paste is a strong data augmentation method for instance
  segmentation.
CoRR
\textbf{abs/2012.07177}
(2020)
\end{botherref}
\endbibitem

\bibitem{DBLP:journals/corr/RosSAW16}
\begin{botherref}
\oauthor{\bsnm{Ros}, \binits{G.}},
\oauthor{\bsnm{Stent}, \binits{S.}},
\oauthor{\bsnm{Alcantarilla}, \binits{P.F.}},
\oauthor{\bsnm{Watanabe}, \binits{T.}}:
Training constrained deconvolutional networks for road scene semantic
  segmentation.
CoRR
\textbf{abs/1604.01545}
(2016)
\end{botherref}
\endbibitem

\bibitem{8954111}
\begin{bchapter}
\bauthor{\bsnm{Li}, \binits{Y.}},
\bauthor{\bsnm{Vasconcelos}, \binits{N.}}:
\bctitle{Efficient multi-domain learning by covariance normalization}.
In: \bbtitle{2019 IEEE/CVF Conference on Computer Vision and Pattern
  Recognition (CVPR)},
pp. \bfpage{5419}--\blpage{5428}
(\byear{2019})
\end{bchapter}
\endbibitem

\bibitem{DBLP:conf/nips/Ben-DavidBCP06}
\begin{botherref}
\oauthor{\bsnm{Ben{-}David}, \binits{S.}},
\oauthor{\bsnm{Blitzer}, \binits{J.}},
\oauthor{\bsnm{Crammer}, \binits{K.}},
\oauthor{\bsnm{Pereira}, \binits{F.}}:
Analysis of representations for domain adaptation.
In: Sch{\"{o}}lkopf, B., Platt, J.C., Hofmann, T. (eds.)
Advances in Neural Information Processing Systems 19, Proceedings of the
  Twentieth Annual Conference on Neural Information Processing Systems,
  Vancouver, British Columbia, Canada, December 4-7, 2006,
pp. 137--144
\end{botherref}
\endbibitem

\bibitem{DBLP:journals/tmi/WangYYFH19}
\begin{barticle}
\bauthor{\bsnm{Wang}, \binits{S.}},
\bauthor{\bsnm{Yu}, \binits{L.}},
\bauthor{\bsnm{Yang}, \binits{X.}},
\bauthor{\bsnm{Fu}, \binits{C.}},
\bauthor{\bsnm{Heng}, \binits{P.}}:
\batitle{Patch-based output space adversarial learning for joint optic disc and
  cup segmentation}.
\bjtitle{{IEEE} Trans. Medical Imaging}
\bvolume{38}(\bissue{11}),
\bfpage{2485}--\blpage{2495}
(\byear{2019})
\end{barticle}
\endbibitem

\bibitem{https://doi.org/10.48550/arxiv.1907.02893}
\begin{botherref}
\oauthor{\bsnm{Arjovsky}, \binits{M.}},
\oauthor{\bsnm{Bottou}, \binits{L.}},
\oauthor{\bsnm{Gulrajani}, \binits{I.}},
\oauthor{\bsnm{Lopez-Paz}, \binits{D.}}:
Invariant Risk Minimization.
arXiv
(2019)
\end{botherref}
\endbibitem

\bibitem{DBLP:conf/icml/AhujaSVD20}
\begin{botherref}
\oauthor{\bsnm{Ahuja}, \binits{K.}},
\oauthor{\bsnm{Shanmugam}, \binits{K.}},
\oauthor{\bsnm{Varshney}, \binits{K.R.}},
\oauthor{\bsnm{Dhurandhar}, \binits{A.}}:
Invariant risk minimization games.
In: Proceedings of the 37th International Conference on Machine Learning,
  {ICML} 2020, 13-18 July 2020, Virtual Event.
Proceedings of Machine Learning Research,
vol. 119,
pp. 145--155
\end{botherref}
\endbibitem

\bibitem{DBLP:conf/icml/KruegerCJ0BZPC21}
\begin{botherref}
\oauthor{\bsnm{Krueger}, \binits{D.}},
\oauthor{\bsnm{Caballero}, \binits{E.}},
\oauthor{\bsnm{Jacobsen}, \binits{J.}},
\oauthor{\bsnm{Zhang}, \binits{A.}},
\oauthor{\bsnm{Binas}, \binits{J.}},
\oauthor{\bsnm{Zhang}, \binits{D.}},
\oauthor{\bsnm{Priol}, \binits{R.L.}},
\oauthor{\bsnm{Courville}, \binits{A.C.}}:
Out-of-distribution generalization via risk extrapolation (rex).
In: Meila, M., Zhang, T. (eds.)
Proceedings of the 38th International Conference on Machine Learning, {ICML}
  2021, 18-24 July 2021, Virtual Event.
Proceedings of Machine Learning Research,
vol. 139,
pp. 5815--5826
\end{botherref}
\endbibitem

\bibitem{DBLP:conf/aaai/LiSWZRLK022}
\begin{botherref}
\oauthor{\bsnm{Li}, \binits{B.}},
\oauthor{\bsnm{Shen}, \binits{Y.}},
\oauthor{\bsnm{Wang}, \binits{Y.}},
\oauthor{\bsnm{Zhu}, \binits{W.}},
\oauthor{\bsnm{Reed}, \binits{C.}},
\oauthor{\bsnm{Li}, \binits{D.}},
\oauthor{\bsnm{Keutzer}, \binits{K.}},
\oauthor{\bsnm{Zhao}, \binits{H.}}:
Invariant information bottleneck for domain generalization.
In: Thirty-Sixth {AAAI} Conference on Artificial Intelligence, {AAAI} 2022,
  Thirty-Fourth Conference on Innovative Applications of Artificial
  Intelligence, {IAAI} 2022, The Twelveth Symposium on Educational Advances in
  Artificial Intelligence, {EAAI} 2022 Virtual Event, February 22 - March 1,
  2022,
pp. 7399--7407
\end{botherref}
\endbibitem

\bibitem{zhou2019semantic}
\begin{barticle}
\bauthor{\bsnm{Zhou}, \binits{B.}},
\bauthor{\bsnm{Zhao}, \binits{H.}},
\bauthor{\bsnm{Puig}, \binits{X.}},
\bauthor{\bsnm{Xiao}, \binits{T.}},
\bauthor{\bsnm{Fidler}, \binits{S.}},
\bauthor{\bsnm{Barriuso}, \binits{A.}},
\bauthor{\bsnm{Torralba}, \binits{A.}}:
\batitle{Semantic understanding of scenes through the ade20k dataset}.
\bjtitle{International Journal of Computer Vision}
\bvolume{127}(\bissue{3}),
\bfpage{302}--\blpage{321}
(\byear{2019})
\end{barticle}
\endbibitem

\bibitem{xiao2010sun}
\begin{bchapter}
\bauthor{\bsnm{Xiao}, \binits{J.}},
\bauthor{\bsnm{Hays}, \binits{J.}},
\bauthor{\bsnm{Ehinger}, \binits{K.A.}},
\bauthor{\bsnm{Oliva}, \binits{A.}},
\bauthor{\bsnm{Torralba}, \binits{A.}}:
\bctitle{Sun database: Large-scale scene recognition from abbey to zoo}.
In: \bbtitle{2010 IEEE Computer Society Conference on Computer Vision and
  Pattern Recognition},
pp. \bfpage{3485}--\blpage{3492}
(\byear{2010}).
\bcomment{IEEE}
\end{bchapter}
\endbibitem

\bibitem{zhou2014learning}
\begin{botherref}
\oauthor{\bsnm{Zhou}, \binits{B.}},
\oauthor{\bsnm{Lapedriza}, \binits{A.}},
\oauthor{\bsnm{Xiao}, \binits{J.}},
\oauthor{\bsnm{Torralba}, \binits{A.}},
\oauthor{\bsnm{Oliva}, \binits{A.}}:
Learning deep features for scene recognition using places database.
Advances in neural information processing systems
\textbf{27}
(2014)
\end{botherref}
\endbibitem

\bibitem{cordts2016cityscapes}
\begin{bchapter}
\bauthor{\bsnm{Cordts}, \binits{M.}},
\bauthor{\bsnm{Omran}, \binits{M.}},
\bauthor{\bsnm{Ramos}, \binits{S.}},
\bauthor{\bsnm{Rehfeld}, \binits{T.}},
\bauthor{\bsnm{Enzweiler}, \binits{M.}},
\bauthor{\bsnm{Benenson}, \binits{R.}},
\bauthor{\bsnm{Franke}, \binits{U.}},
\bauthor{\bsnm{Roth}, \binits{S.}},
\bauthor{\bsnm{Schiele}, \binits{B.}}:
\bctitle{The cityscapes dataset for semantic urban scene understanding}.
In: \bbtitle{Proceedings of the IEEE Conference on Computer Vision and Pattern
  Recognition},
pp. \bfpage{3213}--\blpage{3223}
(\byear{2016})
\end{bchapter}
\endbibitem

\bibitem{neuhold2017mapillary}
\begin{bchapter}
\bauthor{\bsnm{Neuhold}, \binits{G.}},
\bauthor{\bsnm{Ollmann}, \binits{T.}},
\bauthor{\bsnm{Rota~Bulo}, \binits{S.}},
\bauthor{\bsnm{Kontschieder}, \binits{P.}}:
\bctitle{The mapillary vistas dataset for semantic understanding of street
  scenes}.
In: \bbtitle{Proceedings of the IEEE International Conference on Computer
  Vision},
pp. \bfpage{4990}--\blpage{4999}
(\byear{2017})
\end{bchapter}
\endbibitem

\bibitem{dai2017scannet}
\begin{bchapter}
\bauthor{\bsnm{Dai}, \binits{A.}},
\bauthor{\bsnm{Chang}, \binits{A.X.}},
\bauthor{\bsnm{Savva}, \binits{M.}},
\bauthor{\bsnm{Halber}, \binits{M.}},
\bauthor{\bsnm{Funkhouser}, \binits{T.}},
\bauthor{\bsnm{Nie{\ss}ner}, \binits{M.}}:
\bctitle{Scannet: Richly-annotated 3d reconstructions of indoor scenes}.
In: \bbtitle{Proceedings of the IEEE Conference on Computer Vision and Pattern
  Recognition},
pp. \bfpage{5828}--\blpage{5839}
(\byear{2017})
\end{bchapter}
\endbibitem

\bibitem{richter2017playing}
\begin{bchapter}
\bauthor{\bsnm{Richter}, \binits{S.R.}},
\bauthor{\bsnm{Hayder}, \binits{Z.}},
\bauthor{\bsnm{Koltun}, \binits{V.}}:
\bctitle{Playing for benchmarks}.
In: \bbtitle{Proceedings of the IEEE International Conference on Computer
  Vision},
pp. \bfpage{2213}--\blpage{2222}
(\byear{2017})
\end{bchapter}
\endbibitem

\bibitem{zendel2018wilddash}
\begin{bchapter}
\bauthor{\bsnm{Zendel}, \binits{O.}},
\bauthor{\bsnm{Honauer}, \binits{K.}},
\bauthor{\bsnm{Murschitz}, \binits{M.}},
\bauthor{\bsnm{Steininger}, \binits{D.}},
\bauthor{\bsnm{Dominguez}, \binits{G.F.}}:
\bctitle{Wilddash-creating hazard-aware benchmarks}.
In: \bbtitle{Proceedings of the European Conference on Computer Vision (ECCV)},
pp. \bfpage{402}--\blpage{416}
(\byear{2018})
\end{bchapter}
\endbibitem

\bibitem{bevandic2020multi}
\begin{botherref}
\oauthor{\bsnm{Bevandi{\'c}}, \binits{P.}},
\oauthor{\bsnm{Or{\v{s}}i{\'c}}, \binits{M.}},
\oauthor{\bsnm{Grubi{\v{s}}i{\'c}}, \binits{I.}},
\oauthor{\bsnm{{\v{S}}ari{\'c}}, \binits{J.}},
\oauthor{\bsnm{{\v{S}}egvi{\'c}}, \binits{S.}}:
Multi-domain semantic segmentation with pyramidal fusion.
arXiv preprint arXiv:2009.01636
(2020)
\end{botherref}
\endbibitem

\bibitem{2017Accurate}
\begin{botherref}
\oauthor{\bsnm{Goyal}, \binits{P.}},
\oauthor{\bsnm{Dollár}, \binits{P.}},
\oauthor{\bsnm{Girshick}, \binits{R.}},
\oauthor{\bsnm{Noordhuis}, \binits{P.}},
\oauthor{\bsnm{Wesolowski}, \binits{L.}},
\oauthor{\bsnm{Kyrola}, \binits{A.}},
\oauthor{\bsnm{Tulloch}, \binits{A.}},
\oauthor{\bsnm{Jia}, \binits{Y.}},
\oauthor{\bsnm{He}, \binits{K.}}:
Accurate, large minibatch sgd: Training imagenet in 1 hour
(2017)
\end{botherref}
\endbibitem

\bibitem{DBLP:conf/eccv/LinMBHPRDZ14}
\begin{botherref}
\oauthor{\bsnm{Lin}, \binits{T.}},
\oauthor{\bsnm{Maire}, \binits{M.}},
\oauthor{\bsnm{Belongie}, \binits{S.J.}},
\oauthor{\bsnm{Hays}, \binits{J.}},
\oauthor{\bsnm{Perona}, \binits{P.}},
\oauthor{\bsnm{Ramanan}, \binits{D.}},
\oauthor{\bsnm{Doll{\'{a}}r}, \binits{P.}},
\oauthor{\bsnm{Zitnick}, \binits{C.L.}}:
Microsoft {COCO:} common objects in context.
In: Fleet, D.J., Pajdla, T., Schiele, B., Tuytelaars, T. (eds.)
Computer Vision - {ECCV} 2014 - 13th European Conference, Zurich, Switzerland,
  September 6-12, 2014, Proceedings, Part {V}.
Lecture Notes in Computer Science,
vol. 8693,
pp. 740--755
\end{botherref}
\endbibitem

\bibitem{DBLP:journals/corr/abs-1805-04687}
\begin{botherref}
\oauthor{\bsnm{Yu}, \binits{F.}},
\oauthor{\bsnm{Xian}, \binits{W.}},
\oauthor{\bsnm{Chen}, \binits{Y.}},
\oauthor{\bsnm{Liu}, \binits{F.}},
\oauthor{\bsnm{Liao}, \binits{M.}},
\oauthor{\bsnm{Madhavan}, \binits{V.}},
\oauthor{\bsnm{Darrell}, \binits{T.}}:
{BDD100K:} {A} diverse driving video database with scalable annotation tooling.
CoRR
\textbf{abs/1805.04687}
(2018)
\end{botherref}
\endbibitem

\end{thebibliography}


\end{document}